\documentclass{ecai} 

\usepackage{latexsym}
\usepackage{amssymb}
\usepackage{amsmath}
\usepackage{amsthm}
\usepackage{booktabs}
\usepackage{enumitem}
\usepackage{graphicx}
\usepackage{color}

\newtheorem{theorem}{Theorem}

\newtheorem{proposition}[theorem]{Proposition}

\newcommand{\BibTeX}{B\kern-.05em{\sc i\kern-.025em b}\kern-.08em\TeX}

\usepackage{xcolor}
\usepackage{xspace}
\usepackage{comment}
\usepackage{nicefrac}
\usepackage[ruled,vlined]{algorithm2e}
\usepackage{booktabs}
\usepackage{color} 
\definecolor{RoyalBlue}{cmyk}{1, 0.50, 0, 0}
\definecolor{BeamerBlue}{rgb}{0.2, 0.202, 0.698}
\definecolor{ForestGreen}{cmyk}{0.864, 0.0, 0.429, 0.396}
\definecolor{Brown}{cmyk}{0.0,0.692,0.925,0.529}
\definecolor{MyGreen}{cmyk}{0.95, 0.05, 0.95, 0.05}
\definecolor{MyPurple1}{rgb}{0.45,0.353,0.963}
\definecolor{MyPurple2}{rgb}{0.63,0.4,0.63}
\definecolor{MyYellow}{rgb}{0.901,0.547,0.0}
\definecolor{MyStylishGreen}{rgb}{0.328,0.601,0.169}
\usepackage{wrapfig}
\newif\ifcomments
\commentstrue

\usepackage{amsfonts,amssymb,amsmath}

\newcommand{\Reals}{\ensuremath{\mathbb{R}}}
\newcommand{\RR}{\Reals}

\newcommand{\normkind}[2]{\left\lVert#1\right\rVert_{#2}}

\usepackage{bm} 

\newcommand{\innerprod}[2]{\ensuremath{\left\langle\bm{#1}, \bm{#2}\right\rangle}\xspace}
\newcommand{\XX}{\ensuremath{\mathcal{X}}\xspace} 
\newcommand{\YY}{\ensuremath{\mathcal{Y}}\xspace} 

\newcommand{\ourparagraph}[1]{\paragraph{#1.}\xspace}
\usepackage[
   colorlinks%
   ,plainpages=false
   ,hypertexnames=true
   ,naturalnames
   ,hyperindex
   ,citecolor=RoyalBlue
   ,urlcolor=blue
   ,pdfauthor={}
   ,pdftitle={}
   ,pdfsubject={}
]{hyperref}

\begin{document}


\begin{frontmatter}



\title{Dimensionally Reduced Open-World Clustering: DROWCULA}


\author[A, B]{\fnms{Erencem}~\snm{Ozbey}\orcid{0009-0009-9281-4856}\thanks{Corresponding Author. Email: ozbeyerencem@gmail.com}}
\author[B]{\fnms{Dimitrios I.}~\snm{Diochnos}\orcid{0000-0002-2934-606X}} 

\address[A]{Bogazici University, Istanbul, Turkey}
\address[B]{University of Oklahoma, Norman, OK, USA}


\begin{abstract}
Working with annotated data is the cornerstone of supervised learning.  Nevertheless, providing labels to instances is a task that requires significant human effort. Several critical real-world applications make things more complicated because no matter how many labels may have been identified in a task of interest, it could be the case that examples corresponding to novel classes may appear in the future.  Not unsurprisingly, prior work in this, so-called, `open-world' context has focused a lot on semi-supervised approaches.

Focusing on image classification, somehow paradoxically, we propose a fully unsupervised approach to the problem of determining the novel categories in a particular dataset.  Our approach relies on estimating the number of clusters using Vision Transformers, which utilize attention mechanisms to generate vector embeddings.  Furthermore, we incorporate manifold learning techniques to refine these embeddings by exploiting the intrinsic geometry of the data, thereby enhancing the overall image clustering performance.  Overall, we establish new State-of-the-Art results on single-modal clustering and Novel Class Discovery on CIFAR-10, CIFAR-100, ImageNet-100, and Tiny ImageNet.  We do so, both when the number of clusters is known or unknown ahead of time.  The code is available at: 
\url{https://github.com/DROWCULA/DROWCULA}.

\medskip

\noindent\textbf{Keywords:} Clustering, Novel Class Discovery, Dimension Reduction.
\end{abstract}

\end{frontmatter}


%
%
%

\section{Introduction}
Labeled data are the foundation of supervised learning. In the case of classification, such information captures phenomena as observations are \emph{labeled} by human experts in various domains.  
This labeling process requires 
human effort 
and perhaps 
monetary cost too.  
Unsurprisingly, 
a big portion of machine learning research focuses on developing the most performant models for certain datasets, or, to put this differently, on trying to utilize to the largest extent possible the information 
available to a 
learning algorithm in a specific context.

Sometimes, however, this is not the whole story. Many critical applications, e.g., identifying - potentially new - viruses, requires the development of mechanisms that can categorize instances to \emph{novel} classes, leading to the branch of machine learning algorithms that operate in an \emph{open-world} setting~\citep{OW:1,OW:2}.  
A very popular approach for developing such mechanisms, is the use of \emph{semi-supervised learning (SSL)} methods~\citep{Book:Semisupervised:Long,Book:Semisupervised:Short,SSL:surveyHoos}.  Such methods develop a preliminary model using a, relatively small, labeled dataset and try to identify patterns from the unlabeled dataset that can boost the discriminatory performance of the model that is being developed for the particular task.  Furthermore, SSL not only reduces the human effort needed for data labeling, but also increases the capabilities of trained models and 
allows the development of algorithms that operate in an open-world scenario where not all labels are known ahead of time.

Such \emph{novel class discovery (NCD) algorithms} utilize the existing labeled data, and their performance is tested by how accurately they can predict novel classes. Even though NCD has been investigated in the area of SSL, any clustering algorithm can also detect novel classes since no labels are needed for clustering. In this sense NCD can be viewed as a more informed problem than clustering, where some labels are available and some classes are known a priori.

NCD in Open-World SSL setting is useful in some scenarios, nevertheless, we find three drawbacks in these lines of work:
\emph{(i)}
Real-world scenarios ordinarily deal with the emergence of an unknown number of novel classes, 
\emph{(ii)} 
Larger datasets capture a bigger portion of the underlying distribution that governs a particular phenomenon, and typically this also translates to a need for a larger portion of the dataset to be labeled (by humans),
\emph{(iii)}
Despite their unfair advantage, NCD methods fall behind recent clustering algorithms.

\vspace*{-0.2cm}
\begin{figure}[ht]
    \centering
    \includegraphics[width=1\linewidth]{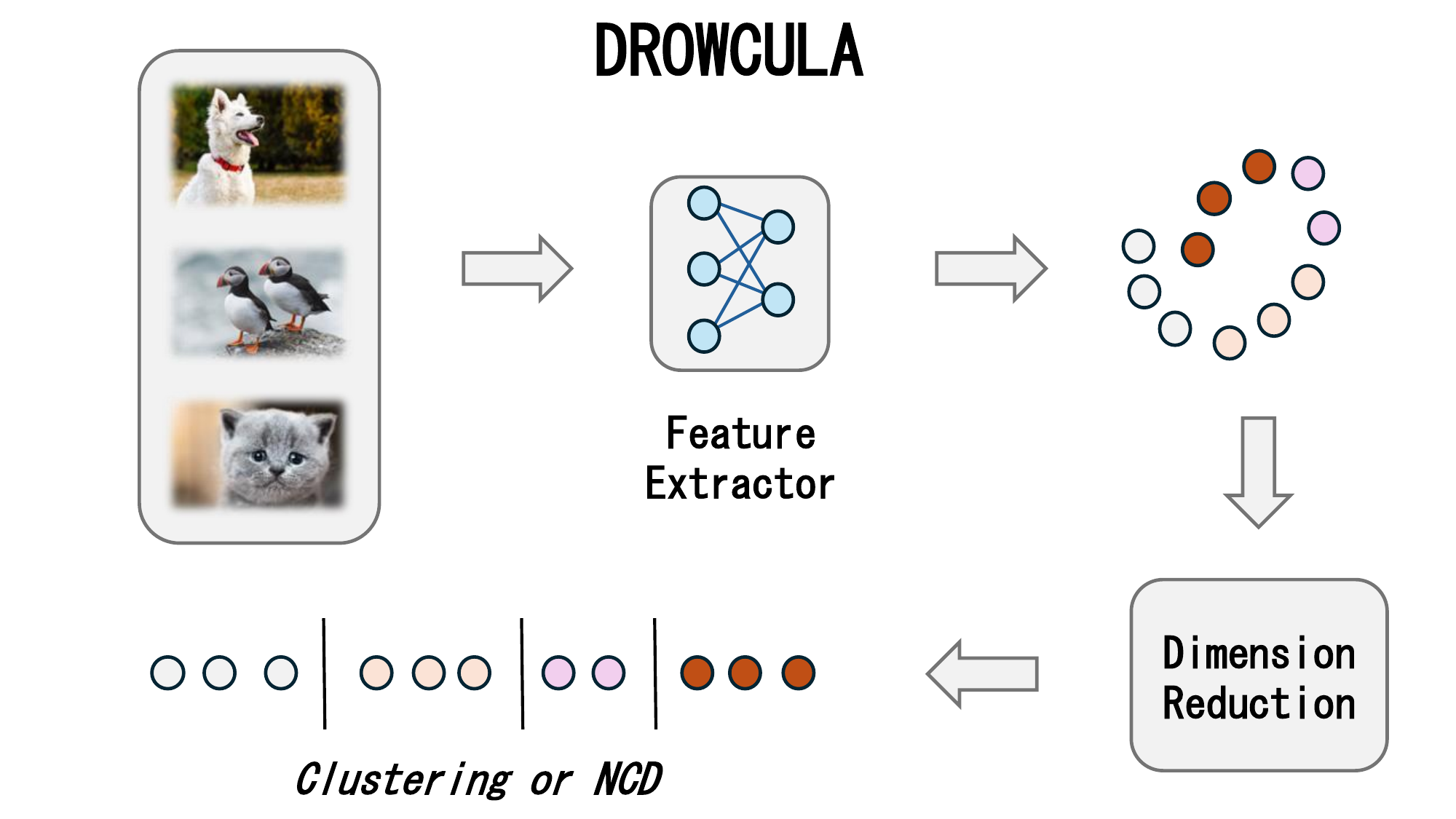}
    \caption{Overview of DROWCULA (our method).
    An image dataset is fed into a feature extraction mechanism and in sequence the dimension of the resulting embeddings is reduced, yielding the low-dimensional instances that we ultimately cluster.}
    \label{fig:drowcula}
    \vspace{1.5em}
\end{figure}

Along these lines, we propose a methodology to completely remove the labeling process for NCD and approach this problem as an area of \emph{Open-World Unsupervised Learning (OWUL)}.
Our proposed methodology \emph{Dimensionally Reduced Open-World Clustering (DROWCULA)} generalizes to large datasets without any prior information and 
outperforms previous clustering and NCD methods. DROWCULA leverages the power of Vision Transformers (ViT) and dimension reduction techniques to create powerful vector embeddings and estimate the number of clusters in the meantime. Figure~\ref{fig:drowcula} provides a schematic overview of our method.

\ourparagraph{\textbf{Contributions}}
Our contributions are summarized as follows. 
First, we propose a \emph{fully unsupervised} setting for 
NCD
in Open-World scenarios.
Second, we establish state-of-the-art (SoTA) accuracies for single-modal clustering and NCD by increasing the efficiency of vector embeddings in categorization tasks.
Third, for clustering in high-dimensional vector spaces, we investigate non-Euclidean metrics and manifold learning methods, while providing a comparison of vision models and clustering algorithms.

\section{Related Work}\label{sec:related}
Image categorization has been an important problem in the field of Computer Vision. Although supervised and unsupervised machine learning algorithms are broadly investigated, open-world settings have received less attention than the development of closed-set machine learning. In an open-world setting, the number of categories and labels are unknown, hence some of the well-accepted classification and clustering methods fall short in these conditions.

Unsupervised learning has been identified as a primary branch of statistical learning for discovering hidden patterns without needing labeled examples. Traditional clustering algorithms 
include 
K-means~\citep{kmeans}, K-medoids~\citep{intruduction_to_cluster_analysis}, Gaussian Mixture Models (GMM) \citep{GMM}, 
density-based methods like DBSCAN \citep{dbscan_revisited2},
and more recently deep clustering approaches~\citep{deepclustering}.
Advancement of pretrained models took the image clustering problem one step further, creating powerful methods such as 
TURTLE, PRCut, PRO-DSC, and TEMI.~\citep{Turtle,PRCut,ProDSC,TEMI}


The main approaches for closed-world 
SSL 
are 
\emph{pseudo-labeling}~\citep{SSL:ST:PL,SSL:MPL,SSL:PL:UncertaintyAware}, 
\emph{consistency regularization}~\citep{Related:SSL:Consistency:1,Related:SSL:Consistency:2},  as well as hybrid 
approaches~\citep{SSL:MixMatch,SSL:ReMixMatch,FixMatch}; see~\citep{SSL:PL:Survey} for a survey and taxonomy of these and related methods.
There is evidence 
that working in an open-world setting is harder than in a closed-world setting, since novel classes among the unlabeled data result in performance loss; e.g.,~\citep{SSL:ClassMismatch}.  
Various approaches have 
tried to deal with this issue; e.g., \citep{SSL:Novel:Solution:DeepSafe,SSL:Novel:Solution:Imbalanced,SSL:Novel:Solution:Robust,ORCA,OpenLDN}. 


While 
SSL
makes a lot of sense in closed-world settings, 
NCD
has been conceived as a subfield of machine learning with an open-world setting in mind, right from the start~\citep{ORCA}.
NCD
relies on the use of a labeled set that bootstraps an initial model that can identify classes, but nevertheless, that learned model has the capability of detecting novel classes~\citep{NCD:Stats,DTC,UNO,NCD:StatsDistillation}. It is therefore not surprising that a large body of work in 
NCD
has focused on open-world 
SSL
approaches, such as~\citet{OpenLDN}.
Acknowledging these 
SSL 
methods, 
our goal 
is 
to discover novel categories in a completely unsupervised manner and, thus, we call this new setting \emph{Open-World Unsupervised Learning (OWUL)}.


Feature extraction and vector embeddings are essential for various computer vision tasks, as they capture the important characteristics of images. 
Deep learning models with Convolutional Neural Network (CNN) architectures had been the standard way of feature extraction; e.g.,~\citep{alexnet,resnet,vgg}. 
%
However, for open-world image clustering or 
NCD,
CNNs become limited due to their dependent nature to the dataset they are trained on. 
On the other hand, Vision Transformers (ViT) have emerged as a promising alternative for these tasks~\cite{Transformers:Images:16x16words,Transformers:Vision:Registers}. ViTs leverage attention mechanisms~\cite{Transformers:Attention}, and their ability to handle large datasets makes them suitable for open-world~\cite{Transformers:Open-World:Detection}.
Utilizing ViTs as foundation models we 
obtain flexible vector representations and improved performance on various tasks.
Hence we propose using ViTs for estimating the number of clusters and performing clustering on entirely unknown datasets.

\smallskip

In what follows, we propose a generalizable OWUL-based methodology leveraging the power of ViTs. Our proposed method \emph{Dimensionally Reduced Open-World Clustering (DROWCULA)} not only eliminates the need for labeled data but also surpasses previously proposed Clustering and NCD methods.

\section{Preliminaries}\label{sec:preliminaries}


\ourparagraph{\textbf{Clustering}}
\label{sec:clustering}
We want to cluster images in similar groups.  However, instead of clustering images directly, we cluster images based on the \emph{embeddings} that we obtain using some vectorization operation on these images.

K-means is one of these clustering algorithms. K-means relies on initializing random centroids, assigning data points to the closest centroid, and updating these centroids by calculating the mean of the assigned data points in that cluster.
%
Different distance metrics result in different K-means clustering performance. Moreover some metrics may be incompatible with the 
representation of the instances; which is true also in some of our scenarios.
%
Instances 
may
be lying on a lower-dimensional manifold, which can deceive Euclidean distances between points and adversely affect the clustering performance. 

%
We investigate clustering algorithms beyond K-means; such as 
K-medoids. 
Instead of using the mean of the data points to determine centroids, it chooses a real data point for this purpose. These center data points are called \emph{medoids}, and they are updated with different heuristic algorithms; Partitioning Around Medoids (PAM)~\citep{pam} is one such 
common implementation. 
The main problem with PAM is the run-time complexity, and for this reason, in practice, one works with optimized algorithms like FastPAM and FasterPAM~\citep{fasterPAM}.

Applying the K-medoids algorithm and then evaluating the Silhouette score is a reasonable methodology; however it is also possible to directly optimize the Silhouette score in the K-medoids algorithm. This technique is known as PAMSIL~\citep{pamsil}; 
with 
run-time optimized versions 
being 
FastMSC and FasterMSC~\cite{FasterMSC}.

\smallskip

In this work, we make our preliminary investigation of distance measures using FasterMSC. Additional details can be found in the Appendix.


\ourparagraph{\textbf{Performance Measures}}
In a totally unlabeled scenario, the performance evaluation of the clustering is not as easy as it is in an image classification problem. To overcome this issue, we use internal performance measures for clustering and external measures for evaluation.
%
External measures do not determine the clustering process, but they are still essential to see the real categorization performance in the end.
We use 
Normalized Mutual Information (NMI), Adjusted Rand Index (ARI), and Clustering Accuracy (ACC) 
as the external measures.
%
As internal measures, we use \emph{Cluster Validity Indices (CVI)}~\citep{cvi_extensive}. Among many options, we chose to use \emph{Silhouette Score (Sil)} for the main clustering evaluation. 
Details and experiments with \emph{Calinski-Harabasz (CHI)}~\citep{calinski-harabasz} - \emph{Davies-Bouldin (DBI)}~\citep{davies-bouldin} indices, and an investigation on the Average Clustering Coefficient ($C_{\text{avg}}$) are included in the Appendix.

\begin{figure}[t]
    \centering
    \includegraphics[width=0.5\columnwidth]{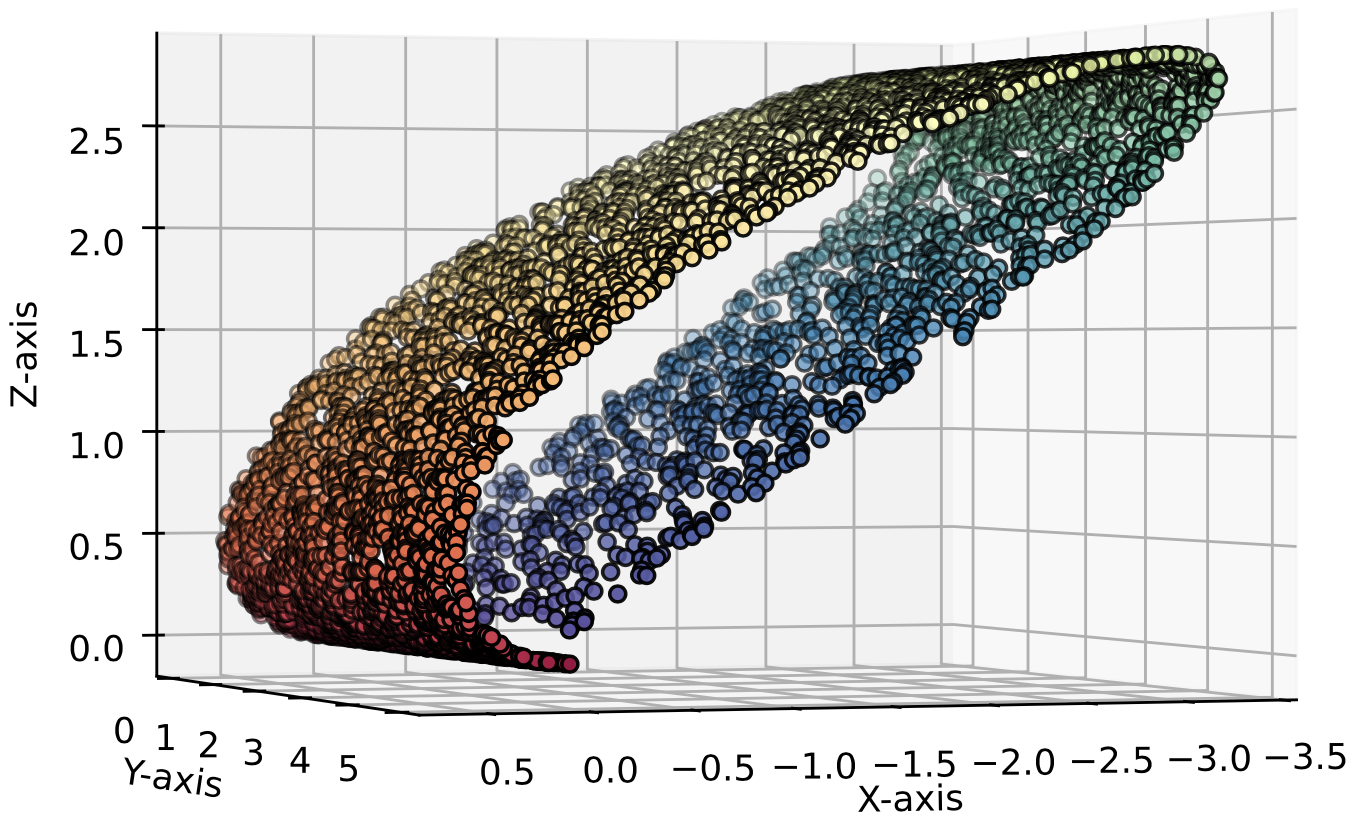}
    \caption{A sample 2D manifold in 3D space. 
    Points that were originally distant in the 2D manifold (e.g., red and blue) 
    end up being close to each other in the 3D space.  Illustration created with artificial data.\\[1.5em]}
    \label{fig:swiss}
\end{figure}

\ourparagraph{\textbf{Dimension Reduction}}\label{sec:dimension-reduction}
In Figure~\ref{fig:swiss} we provide a toy illustration on how treating a lower dimensional manifold in a higher space can cause data points that were originally distant to appear 
closer to each other. 
Case in point are the blue and 
red points in Figure~\ref{fig:swiss}. 
This compression may lead to catastrophic effects for clustering. Hence, we investigate different methods for mapping the data into a 
meaningful lower-dimensional space.
Operating in this space not only improves the clustering quality but also improves the time and space efficiency.

When it comes to dimension reduction, one of the most common approaches is Principal Component Analysis (PCA). PCA finds vectors maximizing the variance of the data. As a linear dimension reduction method, PCA struggles with more complex structures as they cannot be captured linearly.
To address these limitations, we propose apply Manifold Learning methods to preserve the underlying geometry of the data.

Multidimensional Scaling (MDS) is one of these powerful projection methods that preserves the pairwise distance between data points in a lower-dimensional space by applying an eigen decomposition. 
By operating on Geodesic distance local structure of the manifold can be retained. This technique is called Isomap~\cite{isomap}.

Another powerful non-linear method is t-distributed Stochastic Neighbor Embedding (t-SNE)~\cite{t-sne}. This method preserves local similarities by modeling them with a Student-t distribution and minimizing KL divergence between high and low-dimensional representations. While t-SNE is mostly used in data visualization, we believe it is a powerful technique to enhance the performance of clustering. One issue with t-SNE is its computationally expensive nature. As an alternative to t-SNE, Uniform Manifold Approximation and Projection (UMAP)~\cite{umap} can be leveraged.
UMAP is a non-linear manifold learning technique, which constructs a KNN graph and optimizes a similar low-dimensional representation. All of these properties of UMAP make it a strong candidate.

\section{Preliminary Investigation on Clustering}\label{sec:investigation}
In this section we investigate key components of the methodology that we propose in Section~\ref{sec:method}.  Thus, the current section provides a justification of our design choices and can also serve as an ablation study among competing metrics or methods that can be utilized for clustering.

In order to overcome the problematic effect of high-dimensional ViT embeddings, we investigate different distance measures in clustering. Since the K-Means algorithm relies on continuous measures, we preferred to use the FasterMSC algorithm in this experiment.

\ourparagraph{\textbf{Distances}}
In Table \ref{table:distances} we present external performance measurements for the FasterMSC clustering results using different distance measures. Our experiments show that Geodesic distance performs better than 
other measures by overcoming 
\begin{table}[ht]
\caption{Clustering Performance of FasterMSC with Different Distance Measures on CIFAR-10 Dataset.\\[1em]}
\label{table:distances}
\centering
\begin{tabular}{lccc}
\toprule
\textbf{Distance Measure} & \textbf{ACC} & \textbf{NMI} & \textbf{ARI} \\
\midrule
Euclidean & 26.33 & 38.8 & 6.3 \\
Normalized Euclidean & 73.1 & 72.5 & 62.4 \\
Jeffrey Divergence & 70.2 & 69.7 & 53.6 \\
Manhattan & 26.3 & 38.4 & 6.2 \\
Normalized Manhattan & 72.7 & 71.7 & 61.4 \\
Chebyshev & 41.1 & 31.6 & 16.9 \\
Normalized Chebyshev & 56.6 & 39.5 & 34.3 \\

\textbf{Geodesic k=10} & \textbf{83.8} & 
\textbf{88.9} & \textbf{81.9} \\
\hline
\end{tabular}
\end{table}
the curse
of dimensionality. 
Moreover, we observe that FasterMSC is highly sensitive to normalization.

Since we are expected to put similar points in the same cluster, neighboring points in the space should also be clustered into the same category. The superiority of geodesic distance in Table~\ref{table:distances} is related to the prominence of local structure.

\ourparagraph{\textbf{Dimension Reduction}}
Since we need to overcome the curse of dimensionality while conserving the local
\begin{table}[t]
    \caption{Performances of Dimension Reduction Techniques at CIFAR-10 (DINOv2-giant) using K-means.\\[1em]}
    \label{tab:dim_reduction_techniques}
    \centering
    \begin{tabular}{lccc}
        \toprule
        \textbf{Method} & \textbf{ACC} & \textbf{NMI} & \textbf{ARI} \\
        \midrule
        None & 91.0 & 90.8 & 84.3 \\
        Normalization & 91.1 & 90.9 & 84.6 \\
        PCA (30D) & 90.8 & 90.3 & 83.8 \\
        t-SNE (2D) & 98.0 & 95.6 & 95.7 \\
        \textbf{Normalization + t-SNE} & \textbf{98.6} & \textbf{96.6} & \textbf{96.9} \\
        UMAP (3D) & 95.2 & 93.7 & 95.3 \\
        \textbf{Normalization + UMAP} & \textbf{99.1} & \textbf{97.6} & \textbf{98.1} \\
        Autoencoder (16D)& 84.6 & 90.6 & 83.0 \\
        \textbf{Autoencoder + Normalization} & \textbf{99.2} & \textbf{97.6} & \textbf{98.1} \\
        MDS (24D) & 90.7 & 90.1 & 83.5 \\
        \textbf{MDS + Normalization} & \textbf{98.2} & \textbf{95.3} & \textbf{96.0} \\
        Isomap (32D) & 88.4 & 92.5 & 86.8 \\
        \textbf{Isomap + Normalization} & \textbf{97.9} & \textbf{94.6} & \textbf{95.4} \\
        \bottomrule
    \end{tabular}
\end{table}
structure of the data manifold, we have 
evaluated several dimension reduction techniques. After the dimension reduction we obtain an Euclidean space, thus the K-means algorithm is viable in this case.

In Table~\ref{tab:dim_reduction_techniques} we present detailed results of dimension reduction techniques. When compared to Table~\ref{table:distances}, we observe that the K-means algorithm on dimensionally reduced space gives much robust results than FasterMSC on the initial 
space. 
Since we want 
to
obtain a 
transformed space
where the local 
structure is conserved, non-linear dimension reduction methods improve the external clustering measures significantly. Isomap, UMAP, and t-SNE all somehow use neighboring points for dimension reduction, which is essential for clustering tasks. Unlike these non-linear dimension reduction methods, Autoencoders don't directly utilize the intrinsic geometry of the manifold. Hence, vector spaces reduced by Autoencoders are not necessarily well correlated with our evaluation measures as it can be seen in Figure \ref{fig:autoencoder} in the Appendix.

\ourparagraph{\textbf{Feature Extractors}}
In this part, we make a comparison between feature extractors and show the generalizability of the dimension reduction on different models. Our experiments on various pre-trained feature extractors demonstrate that applying dimension reduction can drastically improve the clustering.

\begin{figure*}[t]
    \centering
    \includegraphics[width=1\linewidth]{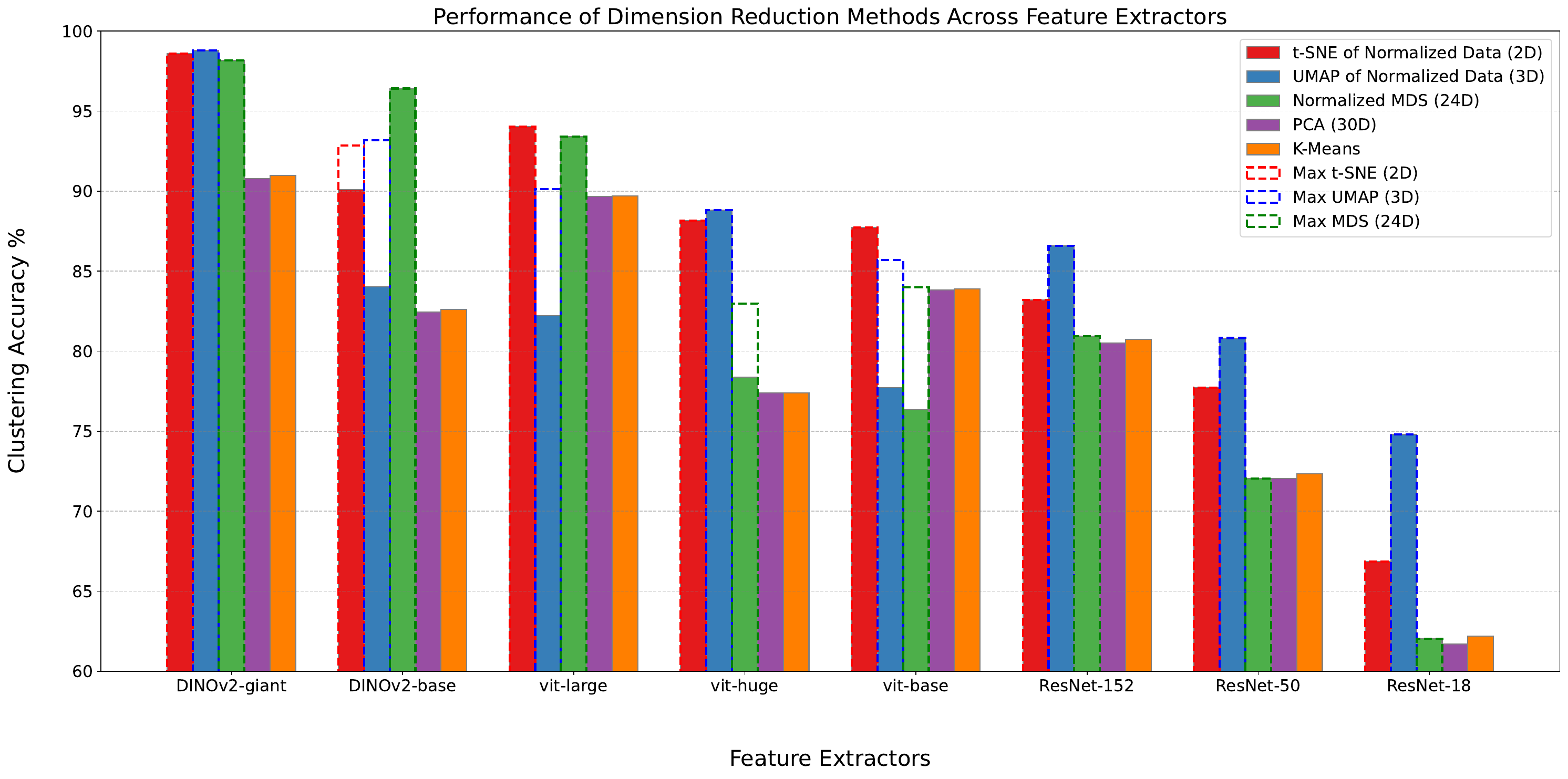}
    \caption{Performance of Feature Extractors on CIFAR-10 Dataset.\\[1em]}
    \label{fig:feature_extractors}
    \vspace{1em}
\end{figure*}

 In Figure \ref{fig:feature_extractors}, we observe that t-SNE (2-D) enhances the performance of the K-means on all feature extractors. Considering DINOv2-giant's 1536-D vectors, our new 2-D vectors are not only better in clustering but also substantially better in memory efficiency.

Our experiments further reveal that 30-D PCA-reduced features are comparable to the original higher-dimensional data in clustering. Although the linear nature of PCA couldn't improve the performance, reducing the number of dimensions saves a significant amount of memory.

Since DINOv2-giant performs better than the other feature extractors among all dimension reduction techniques, we choose to present our proposed methodology using this model. We are aware that most of the pretrained backbones are trained in a supervised/self-supervised way. This latent supervision can also be eliminated by using unsupervised pretrained models.

\section{Methodology}\label{sec:method}

In Open-World scenarios we investigate three main tasks: clustering, determining the number of clusters, and NCD.

\ourparagraph{\textbf{Clustering}}

As we discussed in the previous chapters, it is possible to obtain better performing embeddings with the help of non-linear dimension reduction techniques. In light of our preliminary investigations, we construct our base methodology DROWCULA in the following way.

\begin{itemize}
    \item \textbf{Obtain embeddings:} Use a pretrained Vision Transformer (ViT) to extract feature embeddings from input images.
    
    \item \textbf{L2 normalization:} Apply L2 normalization to each embedding vector.
    
    \item \textbf{Dimensionality reduction:} Reduce the dimensionality of the normalized embeddings using UMAP or t-SNE.
    
    \item \textbf{Clustering:} Apply the K-means algorithm on the reduced embeddings to group them into clusters.
\end{itemize}

To the best of our knowledge, this approach obtains SoTA results in image clustering where vision language models (VLM) are not used adjunctively.


\ourparagraph{\textbf{Number of Cluster Estimation}}
The totally unknown nature of OWUL, makes estimating the number of clusters an essential issue.
In this part, we investigate the usage of DROWCULA in the absence of any prior information.

In Table \ref{tab:dim_reduction_techniques}, we showed that directly using DINOv2 extracted features with the K-Means algorithm can lead to a ACC of 91.0. Although high accuracies can be obtained in the original space, the high dimensional nature of the vector embeddings has a catastrophic effect on internal evaluation measures. 
Applying dimension reduction techniques, results in a significant improvement for these measures; please see Figure~\ref{fig:silhouette_umap} and the related discussion in the Appendix.

Although we expect the best performance at the original number of clusters of the dataset, in some scenarios, overestimation of the number of clusters creates a better clustering performance. Hence, in Figure~\ref{fig:feature_extractors} we also indicate the best clustering performance of our clustering methods with dashed lines to present their real potential in some cases.

This process gives competitive clustering performances compared to knowing the real number of clusters, but computing silhouette scores can be computationally expensive. To maximize Sil in a faster and more robust way, we propose performing \emph{Bayesian Optimization}~\citep{bayesian} on the number of clusters. Experiments on Bayesian Optimization and a pseudo-code for the Number of Cluster Estimation can be found in the Appendix.

\section{Performance Evaluation}\label{sec:evaluation}
In this section, we evaluate DROW\-CULA on clustering - NCD tasks and compare its performance with previous methods.

To the best of our knowledge, Table~\ref{tab:clustering_table} includes the best-performing methods in image clustering. Among these methods, TEMI, PRO-DSC, and TURTLE leverage VLM embeddings (CLIP) while PRCut and DROWCULA only use vision-domain features (DINOv2). 
In this sense, DROWCULA gives state‑of‑the‑art clustering performance in single‑modal vision settings without relying on any language supervision.

\begin{table}[ht]
    \caption{Accuracies of recent clustering algorithms on CIFAR-10 and CIFAR-100 datasets\\[1em]}\label{tab:clustering_table}
    \centering
    \begin{tabular}{l|r|r}
        \toprule
        \textbf{Method}  & \textbf{CIFAR-10} & \textbf{CIFAR-100} \\
        \midrule

        TEMI    \cite{TEMI}     & 96.9  & 73.7 \\
        PRO-DSC \cite{ProDSC}    & 97.2  & 77.3 \\
        
        PRCut-D  \cite{PRCut}   & 79.1 & 80.6 \\

        TURTLE 2-spaces \cite{Turtle} & \textbf{99.5} & \textbf{89.9} \\

        DROWCULA t-SNE  & 98.6 & 81.8   \\
        DROWCULA UMAP   & 99.1 & 80.4 \\

        \bottomrule
    \end{tabular}
\end{table}

In Table~\ref{tab:previous_methods}, we included previous Semi-Supervised NCD and Unsupervised Learning approaches to compare with our proposed methods DROWCULA t-SNE and DROW\-CULA UMAP.

\begin{table}[ht]
    \caption{Accuracy with \textbf{Known} Number of Categories. Novel\% column indicates the percentage of novel categories in the dataset, and subcolumns Known, Novel, and All indicate the clustering accuracy for the corresponding subset of the testing portion of the data. Most of the results from \cite{OpenLDN} and \cite{ORCA}.\\[1em]}
    \label{tab:previous_methods}
    \centering
    \resizebox{\columnwidth}{!}{
    \begin{tabular}{lr|ccc}
        \toprule
        \multicolumn{5}{c}{\textbf{CIFAR-10}} \\
        \textbf{Method} & Novel\% & Known & Novel & All \\
        \midrule
        FixMatch \cite{FixMatch}         & 50\% & 71.5 & 50.4 & 49.5 \\
        DTC \cite{DTC}                   & 50\% & 53.9 & 39.5 & 38.3 \\
        RankStats \cite{NCD:Stats}       & 50\% & 86.6 & 81.0 & 82.9 \\
        UNO \cite{UNO}                   & 50\% & 91.6 & 69.3 & 80.5 \\
        ORCA \cite{ORCA}                 & 50\% & 88.2 & 90.4 & 89.7 \\
        OpenLDN‑MixMatch \cite{OpenLDN}  & 50\% & 95.2 & 92.7 & 94.0 \\
        OpenLDN‑UDA \cite{OpenLDN}       & 50\% & 95.7 & 95.1 & 95.4 \\
        DROWCULA t‑SNE                  & 100\% &    – & 98.6 & 98.6 \\
        DROWCULA UMAP                   & 100\% &    – & \textbf{99.1} & \textbf{99.1} \\
        \midrule
        \multicolumn{5}{c}{\textbf{CIFAR-100}} \\
        \textbf{Method} & Novel\% & Known & Novel & All \\
        \midrule
        FixMatch \cite{FixMatch}         & 50\% & 39.6 & 23.5 & 20.3 \\
        DTC \cite{DTC}                   & 50\% & 31.3 & 22.9 & 18.3 \\
        RankStats \cite{NCD:Stats}       & 50\% & 36.4 & 28.4 & 23.1 \\
        UNO \cite{UNO}                   & 50\% & 68.3 & 36.5 & 51.5 \\
        ORCA \cite{ORCA}                 & 50\% & 66.9 & 43.0 & 48.1 \\
        OpenLDN‑MixMatch \cite{OpenLDN}  & 50\% & 73.5 & 46.8 & 60.1 \\
        OpenLDN‑UDA \cite{OpenLDN}       & 50\% & 74.1 & 44.5 & 59.3 \\
        DROWCULA t‑SNE                  & 100\% &    – & \textbf{81.8} & \textbf{81.8} \\
        DROWCULA UMAP                   & 100\% &    – & 80.4 & 80.4 \\
        \midrule
        \multicolumn{5}{c}{\textbf{ImageNet-100}} \\
        \textbf{Method} & Novel\% & Known & Novel & All \\
        \midrule
        FixMatch \cite{FixMatch}         & 50\% & 65.8 & 36.7 & 34.9 \\
        DTC \cite{DTC}                   & 50\% & 25.6 & 20.8 & 21.3 \\
        RankStats \cite{NCD:Stats}       & 50\% & 47.3 & 28.7 & 40.3 \\
        ORCA \cite{ORCA}                 & 50\% & 89.1 & 72.1 & 77.8 \\
        OpenLDN‑UDA \cite{OpenLDN}       & 50\% & 89.6 & 68.6 & 79.1 \\
        DROWCULA t‑SNE                  & 100\% &    – & \textbf{88.6} & \textbf{88.6} \\
        DROWCULA UMAP                   & 100\% &    – & 83.5 & 83.5 \\
        \midrule
        \multicolumn{5}{c}{\textbf{Tiny ImageNet (200)}} \\
        \textbf{Method} & Novel\% & Known & Novel & All \\
        \midrule
        DTC \cite{DTC}                   & 50\% & 28.8 & 16.3 & 19.9 \\
        RankStats \cite{NCD:Stats}       & 50\% &  5.7 &  5.4 &  3.4 \\
        UNO \cite{UNO}                   & 50\% & 46.5 & 15.7 & 30.3 \\
        OpenLDN‑MixMatch \cite{OpenLDN}  & 50\% & 52.3 & 19.5 & 36.0 \\
        OpenLDN‑UDA \cite{OpenLDN}       & 50\% & 58.3 & 25.5 & 41.9 \\
        DROWCULA t‑SNE                  & 100\% &    – & \textbf{78.4} & \textbf{78.4} \\
        DROWCULA UMAP                   & 100\% &    – & 77.5 & 77.5 \\
        \bottomrule
    \end{tabular}
    }
\end{table}

In these experiments, both of our proposed methods outperformed prior methods in all datasets. Although the difference in ACC between our method and the previous best-performing NCD method 
\begin{table}[ht]
    \caption{Accuracy with \textbf{Unknown} Number of Categories. The columns have similar interpretation as in Table~\ref{tab:previous_methods}.\\[1em]}\label{tab:drowcula_unknown}
    \centering
    \resizebox{\columnwidth}{!}{
    \begin{tabular}{lrccc}
         \toprule
        \textbf{Method} & Novel\% & \multicolumn{3}{c}{\textbf{CIFAR-100}} \\
        & & Known & Novel & All  \\
        \midrule

        DTC \cite{DTC} & 50\% &30.7 & 15.4 & 14.5  \\
        RankStats \cite{NCD:Stats}& 50\% & 33.7 & 22.1 & 20.3 \\
        ORCA \cite{ORCA} & 50\% & 66.3 & 40.0 & 46.4 \\
        DROWCULA t-SNE  &100\% &- & 79.0 & 79.0  \\
        DROWCULA UMAP  &100\% & - & \textbf{80.0} & \textbf{80.0}  \\
        \bottomrule
    \end{tabular}
    }
\end{table}
(OpenLDN-UDA) is less than 4\% on CIFAR-10, our methods outperform OpenLDN-UDA by 74\% on  CIFAR-100, 30\% on ImageNet-100, and by 207\% on Tiny ImageNet in terms of Novel Category accuracy.

Table~\ref{tab:drowcula_unknown} provides a direct comparison between our method and other NCD approaches capable of clustering without prior knowledge of the number of categories. On the CIFAR-100 dataset, our method achieves twice the ACC of ORCA, 2.6 times higher than RankStats, and 4.2 times higher than DTC \citep{ORCA,NCD:Stats,DTC}.

\section{Conclusion and Future Work}\label{sec:conclusion}

Image categorization is an important area of machine learning, and we believe that removing human labeling from image categorization can speed up the development of various related tasks, including the development of foundation models.

In this work we have presented a fully unsupervised methodology for clustering and NCD that leverages ViTs for creating vector embeddings that effectively capture complex features in the data. To address challenges caused by the high-dimensional nature of image embeddings, we employed various dimension reduction methods that not only enhance clustering performance but also improve computational efficiency.

Our comparisons across different datasets demonstrate the superior performance of our approach without having any prior knowledge of the number of clusters. Notably, we establish new SoTA accuracies for single-modal clustering and Novel Class Discovery. We believe that the implications of our work extend beyond clustering, offering a pathway to utilize vast amounts of unlabeled data effectively. By eliminating the human effort, our unsupervised framework may accelerate the development of foundation models and other data greedy tasks in computer vision.

Looking ahead, we intend to further develop our methodology by utilizing other unsupervised learning techniques. Leveraging the features of known datasets, we aim to enhance the quality of our feature representations. We also believe new NCD algorithms can be directly implemented on this setup, thus expanding the literature in a completely new direction in this field.







\bibliography{main}

\newpage
\appendix
\onecolumn

\section{Distances: Metrics and Pseudo-Metrics}

\ourparagraph{Notation}
We use \XX to denote the set of instances and \YY the set of labels. Throughout the paper we use $n$ to denote the dimension of an instance.
For two vectors $\bm{a}, \bm{b}\in\RR^n$, we denote their inner product 
\begin{align}
\innerprod{a}{b} = \bm{a}^T\cdot\bm{b} = \sum_{i=1}^n a_ib_i.
\end{align}

As an alternative to Euclidean distance, we have experimented with Manhattan (L1) Distance, Jeffrey Divergence, Cosine Similarity, and Geodesic Distance.

Manhattan Distance can be useful for computations on high-dimensional data. Its computationally efficient nature makes it a good candidate for large datasets.

On the other hand, cosine-similarity, say for vectors $\bm{a}$ and $\bm{b}$, defined as 
\begin{align}
  \cos(\theta) = \frac{\innerprod{\bm{a}} {\bm{b}}}{\normkind{\bm{a}}{2} \cdot \normkind{\bm{b}}{2}}  
\end{align}
presents a different approach to finding similar data points. 
%
Although cosine similarity is useful, it is linearly related to squared Euclidean distance. For the clustering algorithms that are based on ordering between distances, monotonically transformations, like the squaring operation, are expected to yield similar results. So, the advantages of cosine similarity can also be obtained through L2 normalization over the data.

%

Another distance measure that we evaluate is Jeffreys Divergence. This measure is not considered a true metric since it does not satisfy the triangle inequality, one of the key properties for a metric. Jeffreys Divergence is calculated as the sum of both ways Kullback-Leibler (KL) divergences, where KL divergence measures the difference between two probabilities but is inherently asymmetric. 

\begin{align}
    D_{\text{KL}}(P \parallel Q) = \sum_{x} P(x) \log\left(\frac{P(x)}{Q(x)}\right)
\end{align}

Jeffreys divergence addresses this issue by taking the sum of both directions of KL divergence.

\begin{align}
    J(P \parallel Q) = D_{\text{KL}}(P \parallel Q) + D_{\text{KL}}(Q \parallel P)\,,
\end{align}

Geodesic distance represents the shortest path between two points on a manifold. Since distances are measured along the manifold, Geodesic distance reduces the negative impacts of high-dimensional spaces and overcomes the curse of dimensionality. To calculate this distance, we create a K-Nearest Neighbors (KNN) graph where each point is connected to its closest neighbors. Shortest paths between all pairs of nodes are computed using the Floyd-Warshall algorithm. Since the KNN graph is leveraged, Geodesic distance preserves the local nature of the data manifold.

\section{Number of Cluster Estimation}

As mentioned in the Section~\ref{sec:method}, we provide a pseudo code for the Number of Cluster Estimation in this Section of our Appendix.

Algorithm~\ref{alg:drowcula} explains a generic way of estimating the number of clusters by applying DROWCULA. For simplicity, we define the DINOv2(img) function as a feature extractor and introduce DimReduce(X, method) to apply the chosen dimension reduction method over extracted features. 

\begin{algorithm}[ht]
\caption{Number of Cluster Estimation}\label{alg:drowcula}

\KwData{Image set $Images$; min.\ clusters $k_{\min}=2$; max.\ clusters $k_{\max}$;
        reduction method $method\in\{\mathrm{UMAP},\text{t\text{-}SNE}\}$}
\KwResult{Best label assignment $BestLabels$ and silhouette score $BestSil$}

$F \gets []$\;                  
\For{$img \in Images$}{
    $F \gets F \cup \text{DINOv2}(img)$\;
}

$X \gets \text{Normalize}(F)$\;
$Y \gets \text{DimReduce}(X,\,method)$\;

$BestSil \gets -\infty$\;
$BestLabels \gets \varnothing$\;

\For{$k \gets k_{\min}$ \KwTo $k_{\max}$}{
    $labels \gets \text{KMeans}(Y,\,k)$\;
    $sil \gets \text{Silhouette}(Y,\,labels)$\;
    \If{$sil > BestSil$}{
        $BestSil \gets sil$\;
        $BestLabels \gets labels$\;
    }
}
\Return{$BestLabels,\;BestSil$}\;

\end{algorithm}

\section{Pseudo Labeling}

In semi‑supervised novel‑class discovery (SSL‑NCD), one assumes access to a small set of labeled “seen” classes and a larger pool of unlabeled data that may contain both seen and novel classes. 

In contrast, open‑world unsupervised learning OWUL methods, such as DROWCULA, operate without any true labels. To bring the power of SSL‑NCD into this fully unsupervised setting, we convert DROWCULA’s cluster assignments into pseudo classes that treat each discovered cluster as if it were a “seen” class. 

We obtain these "seen" classes in the following way:
\begin{itemize}
    \item Apply \textbf{DROWCULA} on the dataset.
    \item For each cluster, determine the \textbf{centroid}.
    \item Label the \textbf{closest 50\% of points} to the centroid within each cluster.
\end{itemize}

As shown in Table~\ref{tab:percentile_accuracies}, this simple procedure can give highly accurate pseudo-labels in general. It is also observable that the outer 25\% region gives approximately 20\% worse results than the inner 75\% in CIFAR-100.  These pseudo-labeled instances provided by DROWCULA can therefore be used as starting \emph{labeled instances} in the context of SSL-NCD.

\begin{table}[t]
    \caption{Pseudo Labeling Accuracy (\%) on CIFAR‑10 and CIFAR‑100 at different core‑percentiles using DROWCULA-UMAP.\\\phantom{a}\\[1em]}
    \label{tab:percentile_accuracies}
    \vspace{2 em}
    \centering
    \begin{tabular}{lrr}
        \toprule
        \textbf{Core Percentile} & \textbf{CIFAR‑10} & \textbf{CIFAR‑100} \\
        \midrule
        25\%   & 99.2 & 86.8 \\
        50\%   & 99.3 & 86.3 \\
        75\%   & 99.3 & 85.1 \\
        100\%  & 99.1 & 80.4 \\
        \bottomrule
    \end{tabular}
\end{table}

\section{External Validity Indices}

For two clusterings U and V, NMI gives us an information-theoretic performance metric,  defined as
\begin{align}
NMI(U,V) = \frac{I(U,V)}{\sqrt{H(U) \cdot H(V)}},
\end{align}
where \( I(U,V) \) is mutual information and \( H(U) \), \( H(V) \) are the entropies of the clusterings.
%
ARI gives us a combinatorial way to measure similarity based on the Rand Index ($RI$) and defined as 

\begin{align}
     RI = \frac{(k + l)}{\binom{n}{2}} 
\end{align}

\begin{align}
ARI = \frac{RI - \mathbb{E}[RI]}{Max(RI) - \mathbb{E}[RI]}\,.
\end{align}
In this last case, 
$k$ corresponds to the number of agreed elements between two clusterings
and
$l$ corresponds to the number of disagreed elements between two clusterings.



%

In Section~\ref{sec:preliminaries} we also mention ACC as our main external validity index. 
ACC provides a replacement for the accuracy in the supervised classification. In order to overcome the labeling issue, we used the Hungarian Algorithm to find the optimal assignment for the clusters.

\section{Cluster Validity Indices}

In this section of our Appendix, we provide the effects of dimension reduction on Silhouette (Sil), Calinski-Harabasz (CHI), and Davies-Bouldin (DBI) indices.

Since we are using CVIs to determine the number of clusters, the correlation between these CVIs and our performance evaluation metrics is crucial. Highly correlated results indicate that the maximum/minimum CVIs we obtain also correspond to the maximum/minimum clustering performance

\subsection{Silhouette Score}

Sil is defined as the following:
\begin{align}
\text{Sil} = \frac{1}{N} \sum_{i=1}^{N} \frac{\max\{ d(i), c(i) \} - d(i)}{c(i)}, 
\end{align}


\begin{description}
    \item[$d(i)$] corresponds to the average distance between the point \( i \) and other points in the same cluster.
    \item[$c(i)$] corresponds to the average distance between the point \( i \) and points in the nearest cluster.
\end{description}

\begin{figure}[ht]
    \centering
    \includegraphics[width=1.0\linewidth]{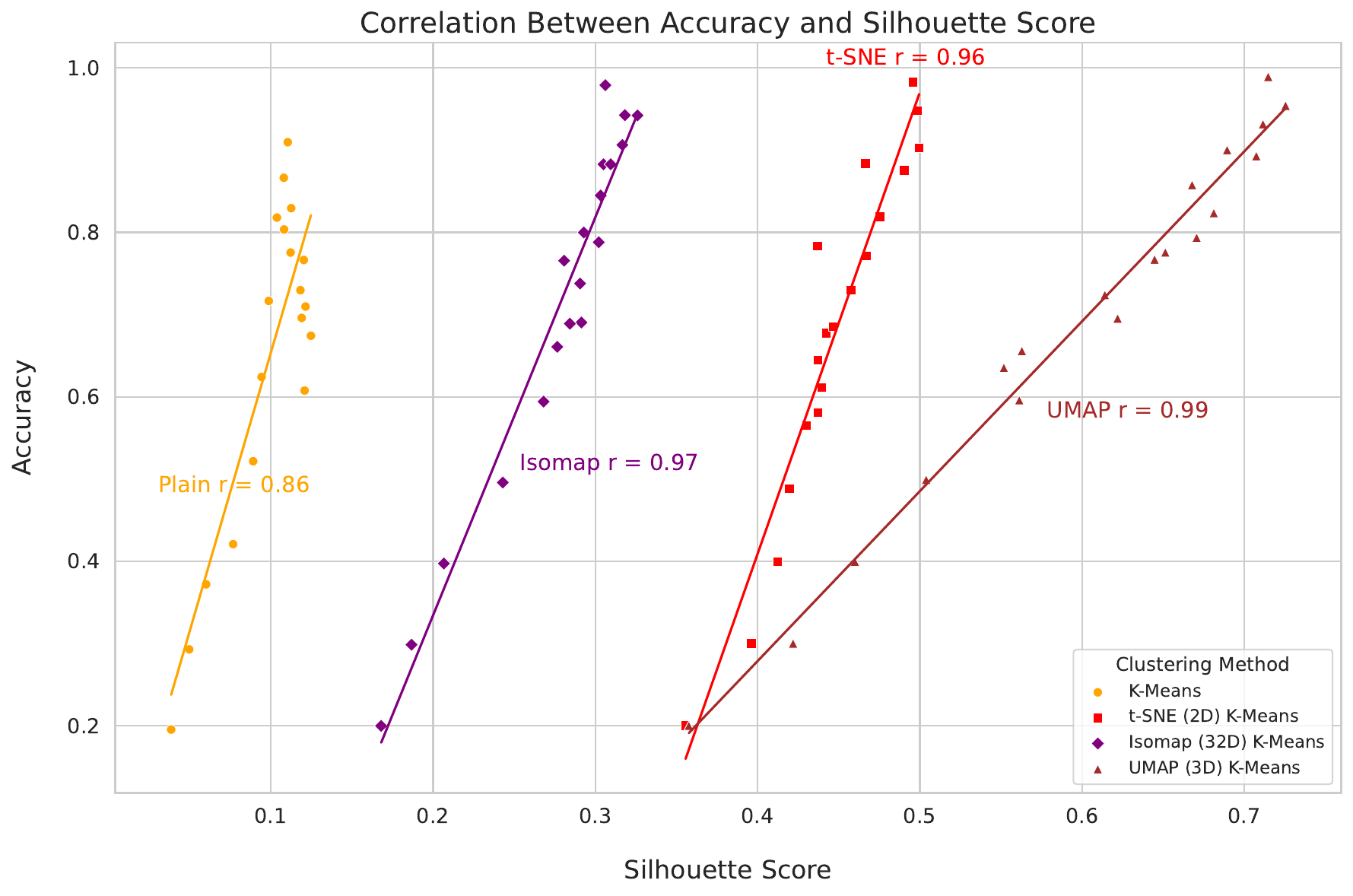}
    \caption{Effect of Dimension Reduction on the Correlation Coefficients between Silhouette Score and Clustering Accuracy on CIFAR-10 Dataset.\\[1em]}
    \label{fig:correlation}
    \vspace*{2em}
\end{figure}

As it can be seen in Figure~\ref{fig:correlation}, investigated dimension reduction techniques increase the correlation coefficient between Sil and Acc by more than 10\% in the CIFAR-10 dataset.

\begin{figure*}[ht]
    \centering
    \includegraphics[width=1\linewidth]{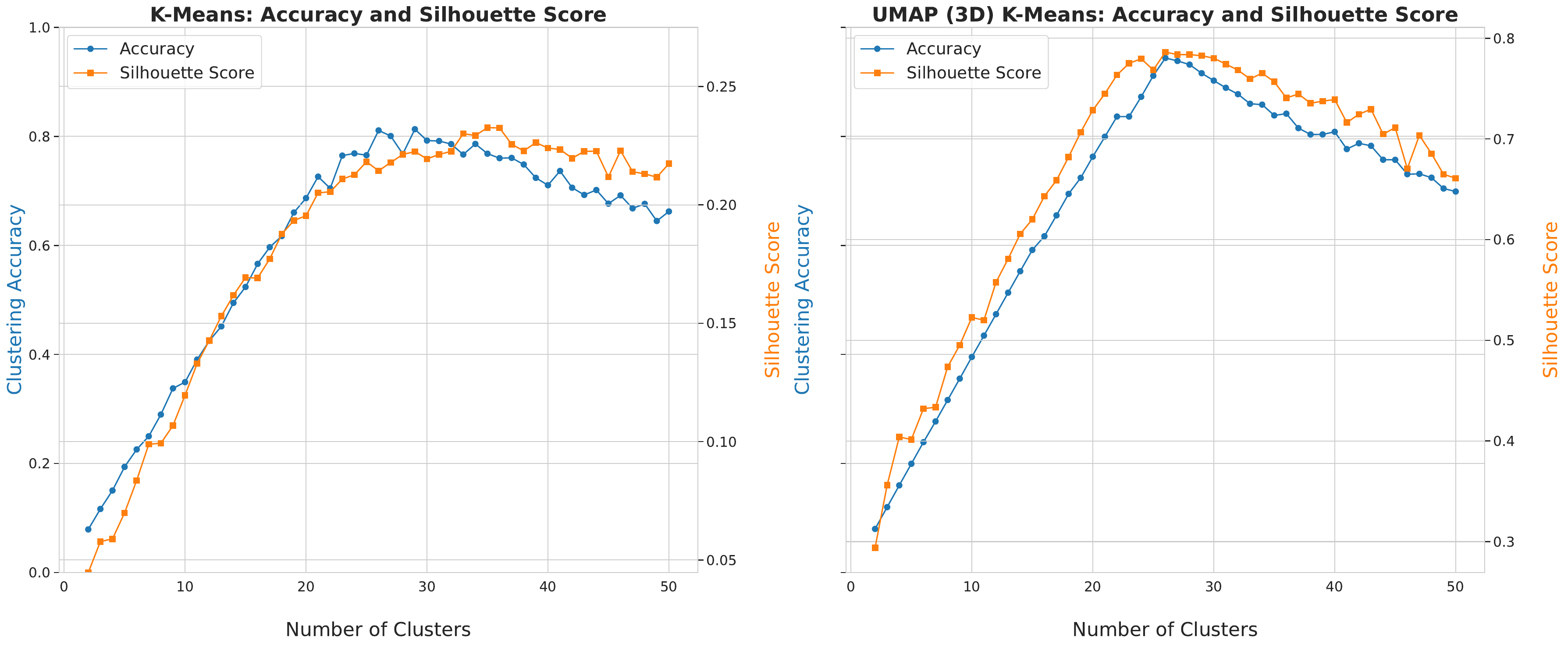}
    \caption{Effect of the UMAP (3D) method on clustering accuracy and silhouette score on a random CIFAR-25 dataset. Acc is marked at the left, and the Sil is marked at the right axis, respectively. Both plots are aligned according to the Acc values, but the Sil is scaled to match with Acc for demonstration purposes.}
    \label{fig:silhouette_umap}
    \vspace{2em}
\end{figure*}

Additionally to what we present in Figure~\ref{fig:correlation}, we further present Figure~\ref{fig:silhouette_umap} to show the performance of our number of cluster estimation method on a 25-cluster subset of the CIFAR-100 dataset. We apply Normalization + UMAP on this subset and compare our results with the K-means algorithm applied to the original vector space. As it can be seen at the right side of Figure~\ref{fig:silhouette_umap}, the highest ACC (94.4\%) is also obtained at the highest Sil value, aligning perfectly with the silhouette score whereas the maximum value of the original silhouette score corresponds to a 76.8\% accuracy.

\subsection{Calinski Harabasz Index}

CHI evaluates clustering quality leveraging two matrices called Between-cluster separation matrix ($B_k$) and Within-cluster dispersion matrix ($W_k$).

\begin{align}
      B_k = \sum_{j=1}^K n_j\,(\mu_j - \mu)\,(\mu_j - \mu)^{\mathsf{T}}.
\end{align}

\begin{align}
    W_k = \sum_{j=1}^K \sum_{x_i \in C_j} (x_i - \mu_j)\,(x_i - \mu_j)^{\mathsf{T}}.
\end{align}

\begin{description}
    \item[$\mu$] overall centroid of all $N$ points.
    \item[$\mu_j$] centroid of cluster $j$.
    \item[$n_j$] number of points in cluster $j$.
    \item[$K$] number of clusters.
\end{description}

Using these two matrices one can calculate CHI as follows:

\begin{align}
\mathrm{CHI} \;=\; \frac{\operatorname{trace}(B_k) \times (N-K)}{\operatorname{trace}(W_k) \times (K - 1)}.
\end{align}

\begin{figure*}[ht]
    \centering
    \includegraphics[width=0.8\linewidth]{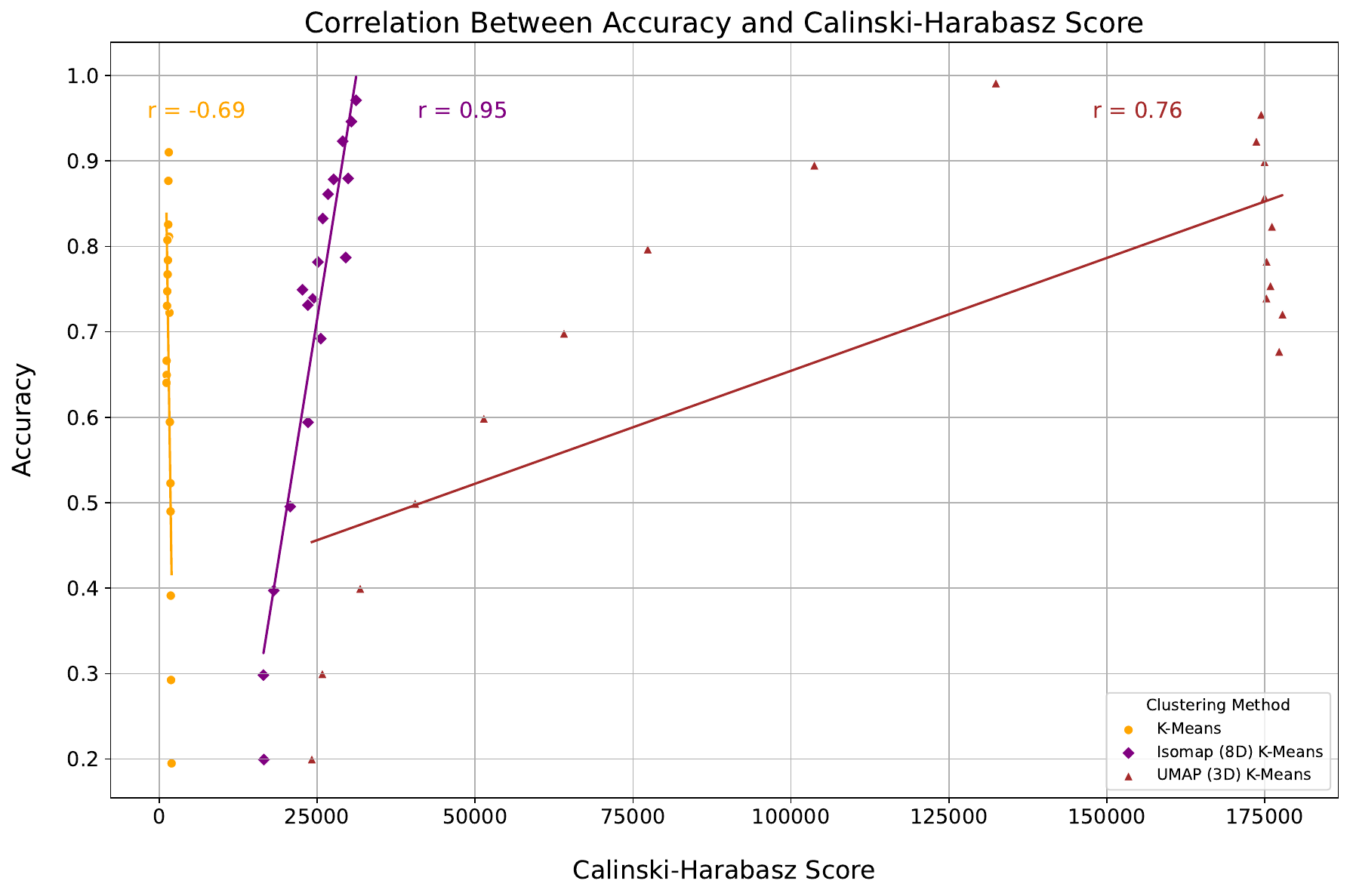}
    \caption{Effect of Dimension Reduction on the Correlation Coefficients between Calinski-Harabasz Index and Clustering Accuracy on CIFAR-10 Dataset.}
    \label{fig:calinski_correlation}
    \vspace{2 em}
\end{figure*}

Figure \ref{fig:calinski_correlation} shows the correlation between CHI and ACC on the CIFAR-10 dataset. The correlation coefficient is -0.69, where there is no dimension reduction technique applied. We observe that Isomap reduction increases this value up to 0.95 ,indicating ACC is highly correlated with CHI.

\begin{figure*}[ht]
    \centering
    \includegraphics[width=1\linewidth]{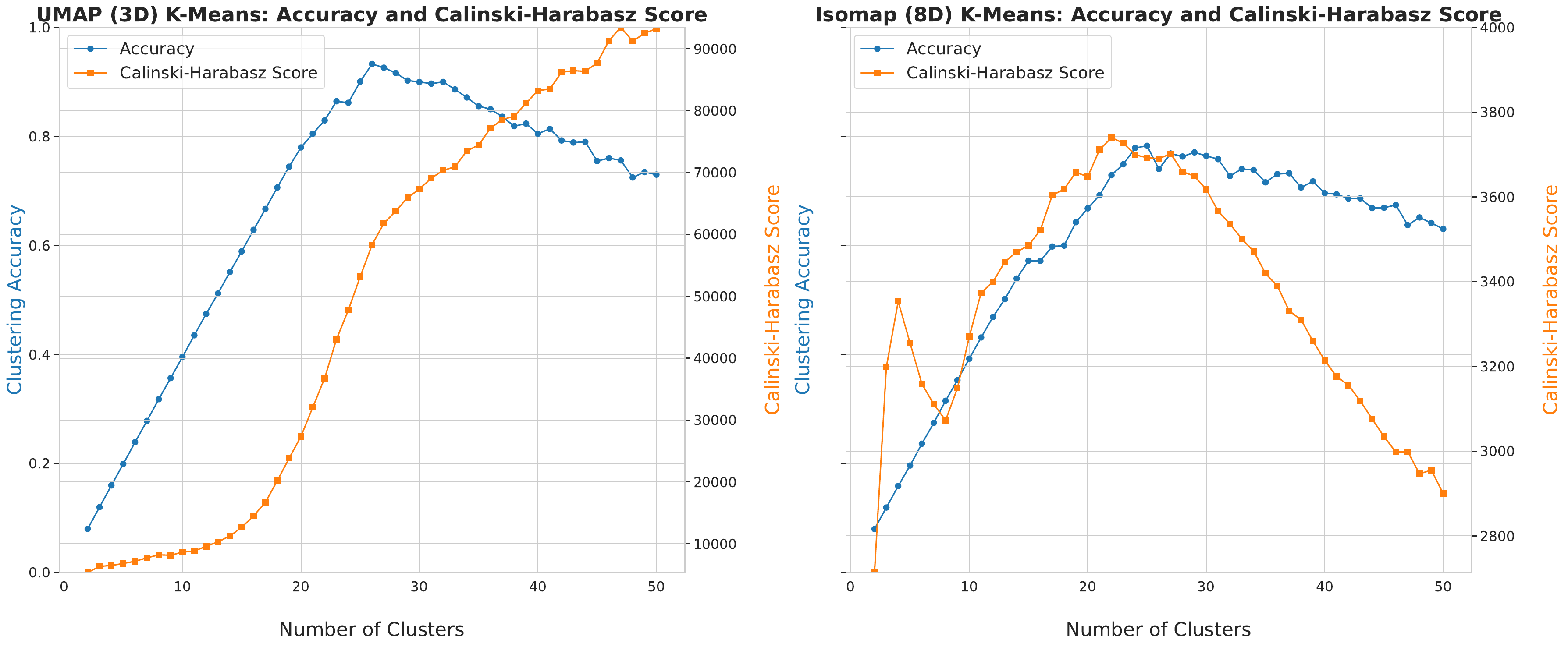}
    \caption{Comparison of the UMAP (3D) and Isomap (8D) methods for Calinski-Harabasz Index (CHI) on a randomly chosen CIFAR-25 dataset.
    Acc is marked at the left, and CHI is marked at the right axis, respectively. Both plots are aligned according to the Acc values, but CHI is scaled to match with Acc for demonstration purposes.\\[1em]}
    \label{fig:calinski_isomap}
\end{figure*}

Despite the fact that UMAP (3D) shows a correlation coefficient of 0.76, the increase in the CHI becomes much smaller after the maximum ACC is observed. Figure~\ref{fig:calinski_isomap} with the CIFAR-25 dataset also supports our observation on Figure~\ref{fig:calinski_correlation}.

\subsection{Davies-Bouldin Index}

DBI evaluates clustering quality via within‐cluster scatter $S_i$ and between‐cluster separation $d_{ij}$.

\begin{align}
    S_i &= \frac{1}{n_i} \sum_{x \in C_i} \lVert x - \mu_i \rVert.
\end{align}


Define the cluster‐pair score:
\begin{align}
    R_{ij} &= \frac{S_i + S_j}{\lVert \mu_i - \mu_j \rVert.}.
\end{align}

Finally, the Davies–Bouldin Index is
\begin{align}
    \mathrm{DBI} &= \frac{1}{K} \sum_{i=1}^{K} \max_{j \neq i} R_{ij}.
\end{align}

\begin{figure*}[ht]
    \centering
    \includegraphics[width=1\linewidth]{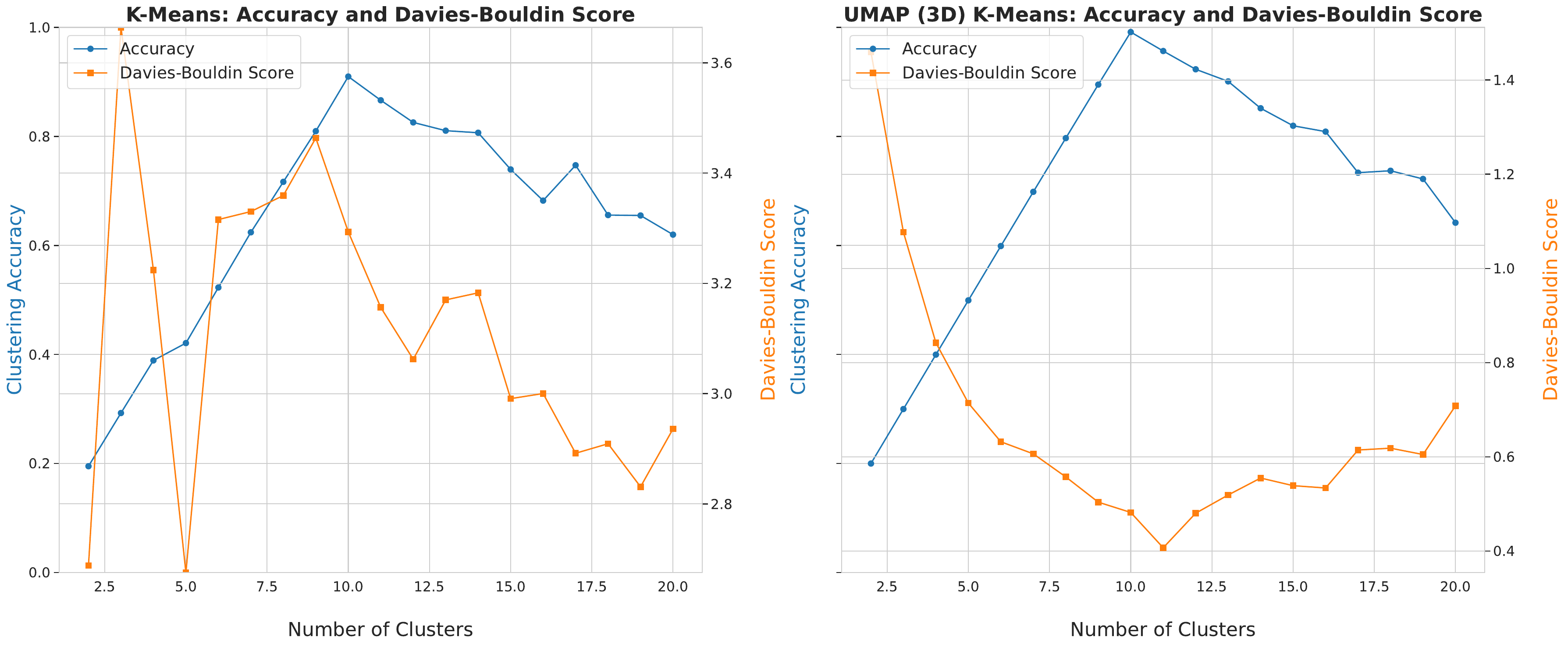}
    \caption{Effect of the UMAP (3D) method on clustering accuracy and Davies-Bouldin Index (DBI) on a random CIFAR-25 dataset. In DBI, smaller values are better.
    Acc is marked at the left, and DBI is marked at the right axis, respectively. Both plots are aligned according to the Acc values, but DBI is scaled to match with Acc for demonstration purposes.\\[1em]}
    \label{fig:dbi_umap}
\end{figure*}

DBI is another well-known CVI exhibiting promising results to determine the number of clusters. Unlike Sil and CHI, as the smaller DBI becomes, clustering gets better. Figure~\ref{fig:dbi_umap} clearly shows that dimension reduction is crucial for DBI. On the original vector space, Sil behaves much better than both CHI and DBI; however, in a reduced vector space, these CVIs can be used in a cooperative manner.

\subsection{Average Clustering Coefficient}

Unlike the CVIs we discussed previously, the clustering coefficient is not related to the clustering process, but it is a graph measure that quantifies the clusterability in a network.
That is, the clustering coefficient is
\begin{displaymath}
C_{\text{avg}} = \frac{1}{N} \sum_{v=1}^{N} C_v\,,
\end{displaymath}
where 
\begin{displaymath}
C_v = \frac{2 \times \text{number of edges between the neighbors of } v}{\text{deg}(v) \times (\text{deg}(v) - 1)}\,.
\end{displaymath}

To test the effects on $C_{\text{avg}}$, we construct a KNN graph using our vector 
embeddings and transform the initial vector space into a network clusterability problem. Although the average clustering coefficient does not directly inform us about the clustering performance, it can be used to discuss the intrinsic geometry of our data manifold and the dimensionality of the vector space. High-dimensional spaces tend to be less clusterable due to the curse of dimensionality.

\begin{table}[ht]
    \centering
    \caption{Clustering Coefficient and Accuracies for Different Dimension Reduction Techniques on CIFAR-10 dataset.\\[1em]}
    \label{tab:clustering_coefficient}
    \vspace*{1.5em}
    \begin{tabular}{lcc}
        \toprule
        \textbf{Dimension Reduction} & \textbf{Average Clustering Coefficient} & \textbf{Accuracy} \\
        \midrule
        None   & 0.2955 & 0.9100 \\
        Normalization + t-SNE (2D) & \textbf{0.608} & \textbf{98.5} \\
        Normalization + UMAP (3D)  & \textbf{0.562} & \textbf{98.8} \\
        PCA (30D)                  & 0.312 & 90.8 \\
        Normalized ISOMAP (32D)    & 0.397 & 97.9 \\
        Normalized MDS (24D)       & 0.306 & 98.2 \\
        \bottomrule
    \end{tabular}    
\end{table}

Average clustering coefficient is not a CVI, but it still carries some information on clusterability. Hence, we also include Table~\ref{tab:clustering_coefficient} in this section.

We made our evaluations by creating KNN graphs with 10 nearest neighbors. Since MDS reduction tries to conserve the original distances between data points, the created KNN graph is ideally the same with the KNN graph created by the original vector space. Thus, for vector spaces reduced with MDS, the average clustering coefficient does not truly represent the clusterability of the data. However, we observe a 100\% increase in the clustering coefficient for Normalization + t-SNE (2D) and 90 \% increase in Normalization + UMAP (3D) algorithms.

\section{FasterMSC}
We preferred to make our evaluations with K-Means algorithm in Table \ref{tab:dim_reduction_techniques} to present comparable results with previous works throughout the paper. In Table~\ref{tab:dim_reduction_fastermsc}, we further present results obtained by the FasterMSC algorithm on CIFAR-10 dataset. In order to ensure robustness, we initialized the algorithm ten times and accepted the clustering with the highest Sil. FasterMSC results are highly affected by the number of dimensions, and normalization process. In the original vector space, normalization resulted in a 178\% increase in ACC. Unlike K-Means algorithm, Normalized PCA reduction increases the performance of FasterMSC by 16\% compared to L2 normalization only. Our proposed Normalization + UMAP reduction ends up with 276\% increase in ACC compared to the original vector space.

\begin{table}[ht]
    \centering
    \caption{Performances of Dimension Reduction Techniques at CIFAR-10 (DINOv2-giant) using \textbf{FasterMSC} Algorithm.\\[1em]}
    \label{tab:dim_reduction_fastermsc}
    \vspace*{1.5em}
    \begin{tabular}{lccc}
        \toprule
        \textbf{Method} & \textbf{ACC} & \textbf{NMI} & \textbf{ARI} \\
        \midrule
        None & 26.3 & 38.8 & 6.3 \\
        Normalization & 73.1 & 72.5 & 62.4 \\
        PCA (30D) & 26.4 & 36.7 & 5.4 \\
        Normalized PCA & 85.1 & 90.6 & 84.1 \\
        \textbf{Normalization + t-SNE} & \textbf{98.5} & \textbf{96.6} & \textbf{96.7} \\
        \textbf{Normalization + UMAP} & \textbf{98.9} & \textbf{97.1} & \textbf{97.6} \\

        MDS (24D) & 32.3 & 44.6 & 8.9 \\
        Normalized MDS & 85.1 & 90.6 & 84.1 \\
        Isomap (32D) & 74.9 & 87.5 & 75.7 \\
        Normalized Isomap & 84.0 & 89.2 & 82.9 \\
        \bottomrule
    \end{tabular}
\end{table}

\begin{figure}[t]
    \centering
    \includegraphics[width=1\linewidth]{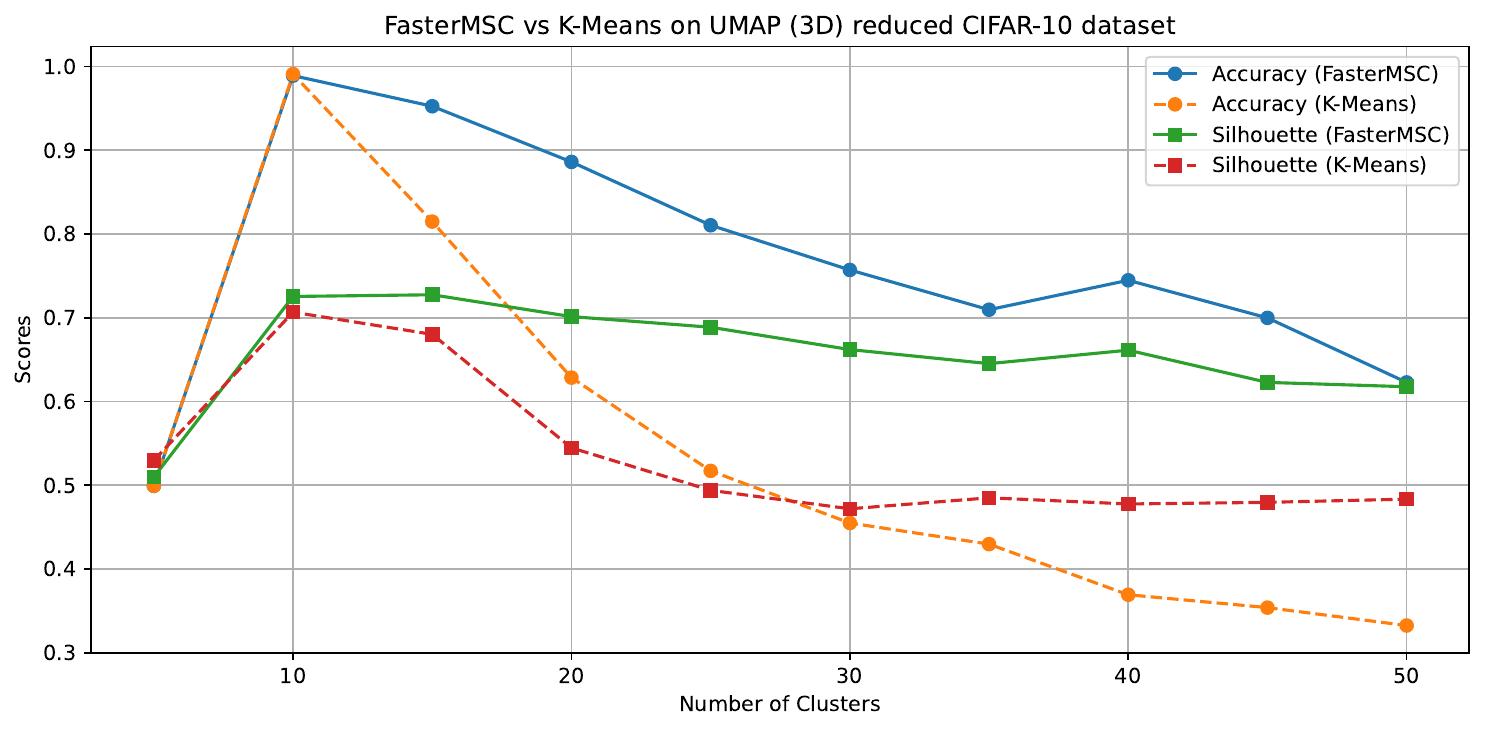}
    \caption{Performance of FasterMSC and K-Means over Normalization + UMAP (3D) applied CIFAR-10 when initialized with different numbers of clusters.\\ \vspace{1em}}
    \label{fig:fast_vs_k}
\end{figure}

K-Means is a more robust algorithm than FasterMSC in high-dimensional spaces, but when dimension reduction is applied, FasterMSC results are highly promising. Since FasterMSC maximizes Sil while clustering, it is less sensitive to overestimation of a number of clusters. As it can be seen in Figure~\ref{fig:fast_vs_k}, both Sil and ACC of FasterMSC are more stable than results obtained by K-Means. FasterMSC loses less than 4\% ACC between 10 and 15 clusters, where this loss is more than 17\% for K-Means. 

\section{Additional Results}
To provide a further comparison between our methodology, we also present ACC scores with an unknown number of categories on these datasets in Table~\ref{tab:drowcula_is_better}. DROWCULA UMAP with an unknown number of categories again performed exemplary better than the prior. Once again, for Tiny ImageNet, the methods presented in Table~\ref{tab:drowcula_is_better} set a new SoTA in clustering ACC.

\begin{table}[ht]
    \caption{Accuracy of DROWCULA with \textbf{Unknown} Number of Categories. The columns have similar interpretation as in Table~\ref{tab:previous_methods}.\\[1em]}
    \label{tab:drowcula_is_better}
    \vspace{2em}
    \centering
    \begin{tabular}{l|rr}
        \toprule
        \textbf{Method}         & \textbf{CIFAR‑10} & \textbf{CIFAR‑100} \\
        \midrule
        DROWCULA t‑SNE          & 90.3              & 79.0               \\
        DROWCULA UMAP           & \textbf{95.4}     & \textbf{80.0}      \\
        \midrule
        \textbf{Method}         & \textbf{ImageNet‑100} & \textbf{Tiny ImageNet (200)} \\
        \midrule
        DROWCULA t‑SNE          & 85.2                  & \textbf{79.4}                \\
        DROWCULA UMAP           & \textbf{83.8}         & 78.4                         \\
        \bottomrule
    \end{tabular}
\end{table}

\subsection{Bayesian Optimization}
 Since we increase the correlation between Sil and ACC as it can be seen in Figure~\ref{fig:correlation}, applying search algorithms to find the maximum value of Sil becomes applicable. 

Table~\ref{tab:bayesian} clearly shows that Bayesian Optimization can speed up the searching process with a relatively small compromise in ACC. Since the number of iterations for Bayesian Optimization is a hyperparameter, the trade-off between ACC and the number of iterations can be externally determined.

\begin{table}[ht]
    \centering
    \caption{Effects of Bayesian Optimization across different datasets using DROWCULA UMAP (3D) on normalized data.\\[1em]}\label{tab:bayesian}
    \vspace{2em}
    \begin{tabular}{lcccc}
        \toprule
        & \multicolumn{2}{c}{CIFAR-10} & \multicolumn{2}{c}{CIFAR-100} \\
        & ACC & \# Iteration & ACC & \# Iteration \\
        \midrule
        Bayesian   & 98.8   & 5 & 75.9 & 20 \\
        Manual     & 98.8   & 20 & 80.0 & 200\\
        \midrule
        & \multicolumn{2}{c}{Imagenet-100} & \multicolumn{2}{c}{Tiny ImageNet} \\
        & ACC & \# Iteration & ACC & \# Iteration \\
        \midrule
        Bayesian & 84.0 & 20  & 76.5  & 40  \\
        Manual   & 83.8 & 200 & 78.4  & 400 \\
        \bottomrule
    \end{tabular}
\end{table}

\begin{figure}[t]
    \centering
    \includegraphics[width=0.8\linewidth]{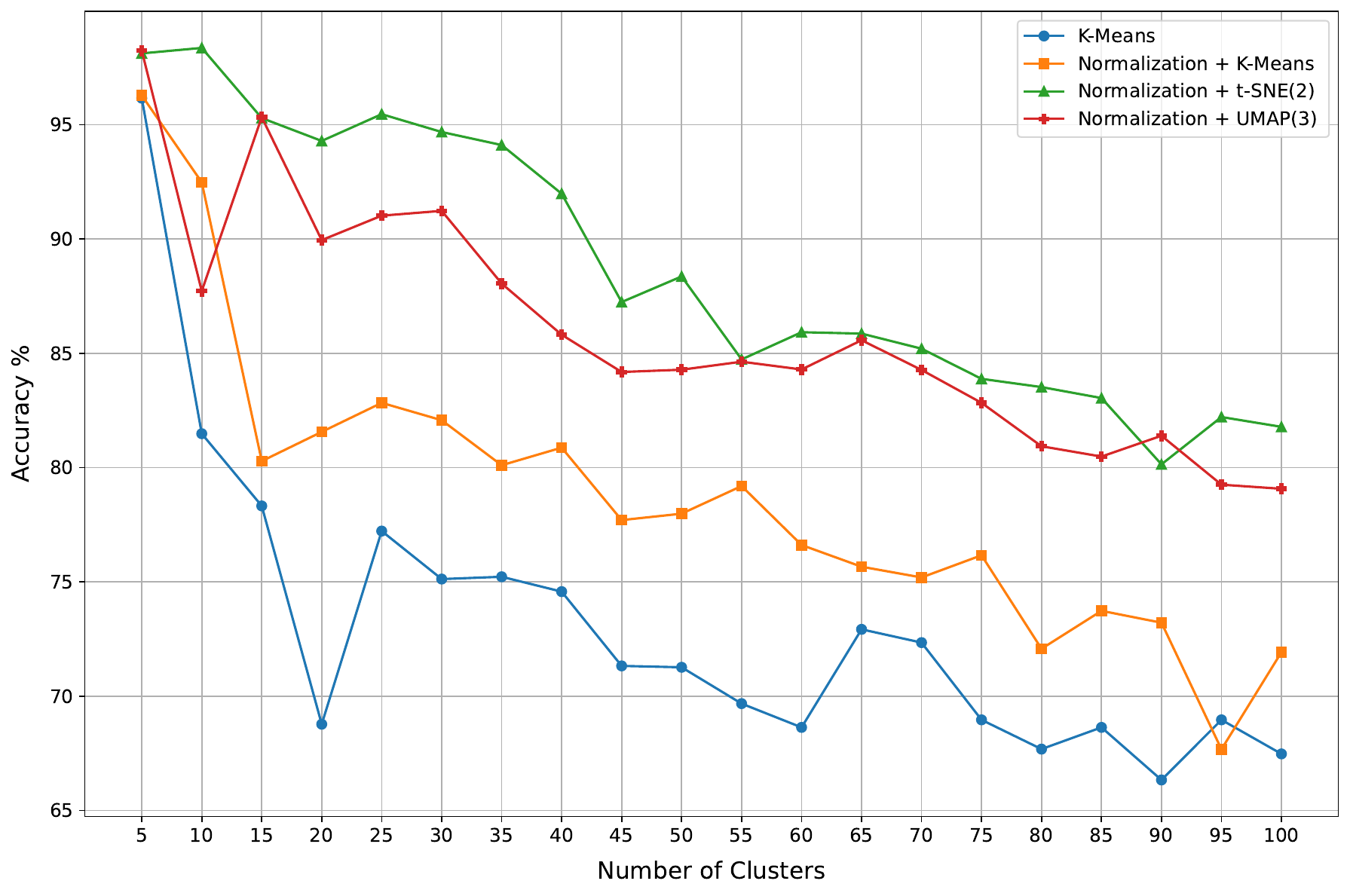}
    \caption{Performance of Dino-v2-giant on some dimension reduction techniques over random subsets of CIFAR-100 having different numbers of clusters.}
    \label{fig:sampled100}
    \vspace*{2.5em}
\end{figure}

In Table~\ref{tab:previous_methods}, we showed the performance of DROWCULA with different datasets. Figure~\ref{fig:sampled100} also provides additional results for random subsets of CIFAR-100. These subsets are chosen without replacement, and larger subsets also include all other smaller subsets (CIFAR-30 is also a subset of CIFAR-35). As expected, ACC decreases with the increasing size of subsets. For all subsets, t-SNE (2D) and UMAP (3D) enhanced the performance of the K-Means. Another important result that can be obtained from Figure~\ref{fig:sampled100} is their response to the increasing number of clusters. Original vector space loses more than 25\% of ACC in the first 20 clusters, whereas this value is smaller than 5\% in t-SNE reduced vector space. In the majority of these subsets, t-SNE also performed better than UMAP reduction. Although some portion of the fluctuation of UMAP can be reduced by trying different k values in K-means (We observe the same aspect in Figure \ref{fig:feature_extractors}), t-SNE can be considered as a more robust dimension reduction technique for CIFAR-100. Still, UMAP is preferable due to its computationally efficient nature and comparable results with t-SNE.

\subsection{Autoencoder Reduction}

In Section~\ref{sec:investigation}, we mentioned that Autoencoders can be inconsistent since they don't directly conserve the local structure of the data manifold. Clustering quality often relies on preserving neighborhood relations. Transformations that distort local structure can yield unstable clustering performance.

\begin{figure}[t]
    \centering
    \includegraphics[width=0.8\linewidth]{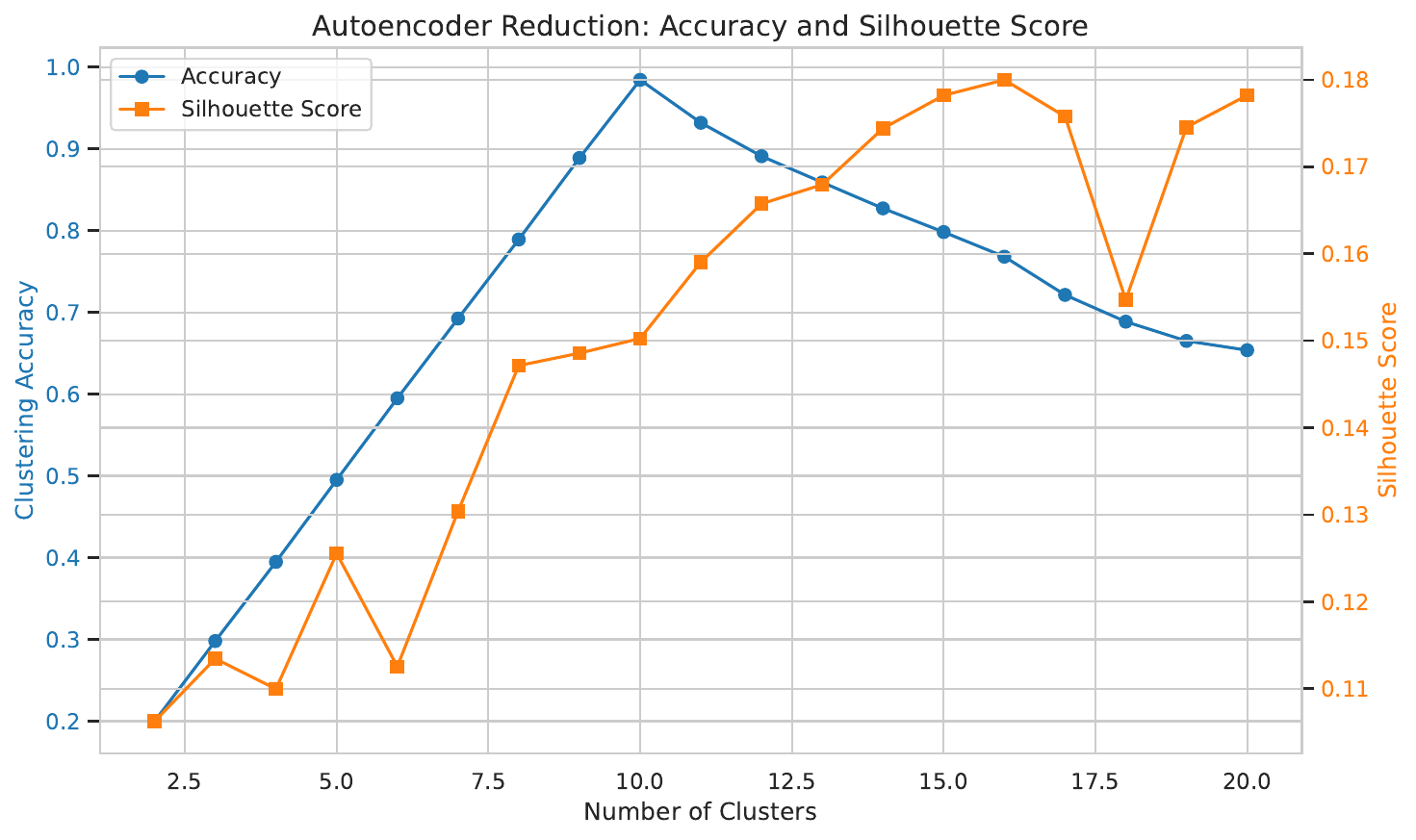}
    \caption{Accuracy and Silhouette Score of DROWCULA-Autoencoder (1536 - 1024 - 256 - 16D) with respect to the number of clusters on CIFAR-10\\[1em]}
    \label{fig:autoencoder}
    \vspace*{2.5em}
\end{figure}

Our experiments showed that, depending on how many hidden layers are used in the Autoencoder, results may fluctuate. To build an example scenario, we created an autoencoder with a 1536 - 1024 - 256 - 16 dimensional architecture. Figure~\ref{fig:autoencoder} also presents the fluctuating nature of autoencoder-based clustering.

\section{Datasets and Experimental Details}

In this chapter of the Appendix, we present datasets and hyperparameters that we used throughout our experiments. 
Details about the datasets that we used are available in Table~\ref{tab:dataset_sizes}.

\begin{table}[ht]
   \caption{Sizes of the datasets used in our evaluation benchmarks.\\[1em]}
   \label{tab:dataset_sizes}
   \vspace{2em}
    \centering
    \begin{tabular}{lccc}
        \toprule
        Dataset & Train & Test & Classes \\
        \midrule
        CIFAR-10 & 50,000 & 10,000 & 10 \\
        CIFAR-100 & 50,000 & 10,000& 100 \\
        ImageNet100 & 126,689& 5,000 & 100 \\
        Tiny ImageNet & 100,000 & 10,000 & 200 \\
        \bottomrule
    \end{tabular}    
\end{table}

\subsection{Datasets}

Our proposed methods, DROWCULA UMAP, and DROWCULA t-SNE, were evaluated on four well-known image categorization datasets. To assess our performance with a small number of novel categories, we made experiments using CIFAR-10 \cite{CIFAR}, for assessment with large numbers of novel categories, we evaluated our methods on CIFAR-100 \cite{CIFAR}, ImageNet-100 \cite{imagenet100,imagenet1000} (a randomly selected subset of ImageNet containing 100 classes \cite{ImageNet}) and Tiny ImageNet \cite{tinyimagenet} respectively. 

To demonstrate the generalizability of our methods to large-scale datasets, we used the training portions of these datasets for evaluation. Clustering in larger datasets presents additional challenges as larger cluster sizes can lead to overlapping or intersecting clusters.

\subsection{Hyperparameters}
In our experiments, we used hyperparameters presented in Table \ref{tab:hyperparameters}. For t-SNE and UMAP, increasing the number of components does not necessarily increase the performance thus, two and three components are chosen for these methods. On the other hand, in PCA, MDS, and Isomap methods increasing the number of dimensions can also end up with better results, hence we chose a large number of components in these scenarios. Only in Figure~\ref{fig:calinski_isomap}, the number of components is chosen as 8 for the Isomap method to obtain a more robust result by decreasing the number of dimensions.

\begin{table}[ht]
    \centering
    \caption{Hyperparameters Used in the Experiments\\[1em]}
    \label{tab:hyperparameters}
    \vspace{2em}
    \begin{tabular}{ll}
        \hline
        \textbf{Method} & \textbf{Hyperparameters} \\
        \hline
        K-Means     & n\_init = 50, max\_iter = 10000 \\
        t-SNE       & n\_components = 2, perplexity = 30, max\_iter = 10000 \\
        UMAP        & n\_components = 3, n\_neighbors = 10, min\_dist = 0.1 \\
        MDS         & n\_components = 24 \\
        Isomap      & n\_components = 8/32 \\
        PCA         & n\_components = 30 \\
        FasterMSC   & max\_iter = 10000 \\
        \hline
    \end{tabular}
\end{table}

\begin{figure}[ht]
    \centering
    \includegraphics[width=1\linewidth]{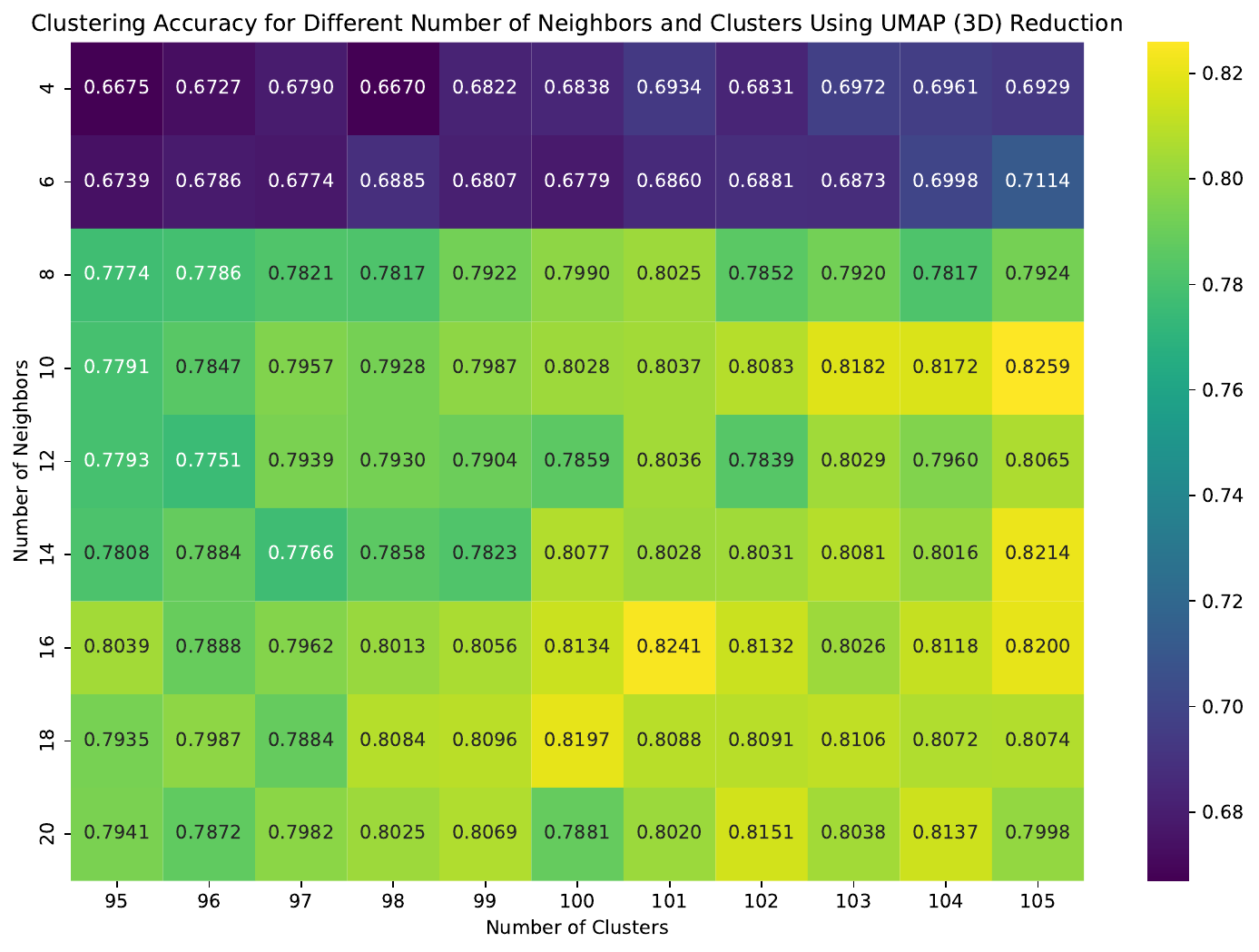}
    \caption{Performance of DROWCULA-UMAP (3D) with respect to the number of clusters and number of neighbors on CIFAR-100 dataset.}
    \label{fig:umap}
\end{figure}

For the UMAP algorithm, we present different ACCs obtained by varying the number of neighbors and number of clusters in Figure \ref{fig:umap}. Our experiments show that having a number of neighbors smaller than 8 performs significantly worse. We recommend keeping this hyperparameter between 10 and 20 to take advantage of local structures in dimension reduction.

\end{document}